\documentclass[final]{cvpr}

\usepackage{times}
\usepackage{epsfig}
\usepackage{graphicx}
\usepackage{amsmath}
\usepackage{amssymb}

\usepackage[pagebackref=true,breaklinks=true,colorlinks,bookmarks=false]{hyperref}

\def\cvprPaperID{6918} % *** Enter the CVPR Paper ID here
\def\confYear{CVPR 2021}
\usepackage{times}
\usepackage{microtype} % microtypography
\usepackage{graphicx} % more modern
\usepackage{url}
\usepackage{subfigure} 
\usepackage{wrapfig}
\usepackage{balance} 
\usepackage{algorithm}
\usepackage{algorithmic}

\renewcommand{\algorithmicrequire}{\textbf{Initialize:}}   
\renewcommand{\algorithmicensure}{\textbf{Iteration:}} 
\renewcommand{\algorithmicinput}{\textbf{Input:}} 
\renewcommand{\algorithmicoutput}{\textbf{Output:}}

\usepackage{hyperref}

\usepackage{helvet}
\usepackage{courier}

\usepackage{makecell}
\usepackage{graphicx}
\usepackage{subfigure}
\usepackage{color}
\usepackage{amsfonts}
\usepackage{amsmath}
\usepackage{amssymb}

\usepackage{multirow}
\usepackage{booktabs}
\usepackage{threeparttable}

\def\ie{\mbox{\textit{i.e.}, }}
\def\eg{\mbox{\textit{e.g.}, }}
\def\wrt{\mbox{\textit{w.r.t. }}}

\def\bkappa{\mbox{{\boldmath $\kappa$}}}

\def\bTheta{\mbox{{\boldmath{$\Theta$}}}}

\def\mA{{\mathcal A}}

\def\mL{{\mathcal L}}

\def\mP{{\mathcal P}}

\def\mS{{\mathcal S}}
\def\mT{{\mathcal T}}

\def\mX{{\mathcal X}}

\DeclareMathAlphabet\mathbfcal{OMS}{cmsy}{b}{n}
\def\0{{\bf 0}}
\def\1{{\bf 1}}

\def\bx{{\bf x}}
\def\by{{\bf y}}
\def\bz{{\bf z}}

\def\mmR{{\mathbb R}}

\def\bx{{\bf x}}

\def\by{{\bf y}}

\def\bz{{\bf z}}

\usepackage{ntheorem}

\newtheorem{deftn}{Definition}
\newtheorem{thm}{Theorem}
\newtheorem*{*thm}{Theorem}

\newtheorem*{*lemma}{Lemma}

\newenvironment*{proof}{\textbf{Proof}\quad}{\hfill $\square$\par}

\usepackage{array}
\usepackage{tabu}
\makeatletter
\newcommand{\nipstophline}{%
	\noalign {\ifnum 0=`}\fi \hrule height 4pt
	\futurelet \reserved@a \@xhline
}
\newcommand{\nipsbottomhline}{%
	\noalign {\ifnum 0=`}\fi \hrule height 1pt
	\futurelet \reserved@a \@xhline
}
\usepackage{enumitem}
\def\blue{\textcolor{blue}}

\usepackage[utf8]{inputenc} % allow utf-8 input
\usepackage[T1]{fontenc}    % use 8-bit T1 fonts
\usepackage{nicefrac}       % compact symbols for 1/2, etc.
\usepackage{wrapfig}

\makeatletter
\renewcommand{\paragraph}{%
  \@startsection{paragraph}{4}%
  {\z@}{1.2ex \@plus 1ex \@minus .2ex}{-1em}%
  {\normalfont\normalsize\bfseries}%
}
\makeatother

\def\ljc{\textcolor{black}}
\def\zsh{\textcolor{black}}
\def\blue{\textcolor{black}}
\def\tmk{\textcolor{black}}

\begin{document}

%%%%%%%%% TITLE
% \title{Counterattacking Adversarial Examples by Defense Transformer}
\title{Learning Defense Transformers for Counterattacking Adversarial Examples}

\author{
    Jincheng Li$^{1}$\thanks{Authors contributed equally.}~, Jiezhang Cao$^{2*}$, Yifan Zhang$^{3}$, Jian Chen$^{1}$, Mingkui Tan$^{1,4}$\thanks{Corresponding author.} \\
    $^{1}$South China University of Technology, $^{2}$ETH Zürich, $^{3}$National University of Singapore \\
    $^{4}$Key Laboratory of Big Data and Intelligent Robot, Ministry of Education \\
    {\tt\small sejinchengli@mail.scut.edu.cn, }
    {\tt\small mingkuitan@scut.edu.cn}
}


\maketitle

\begin{abstract}
	Deep neural networks (DNNs) are vulnerable to adversarial examples with small perturbations. Adversarial defense thus has been an important means which improves the robustness of DNNs by defending against adversarial examples. Existing defense methods focus on some specific types of adversarial examples and may fail to defend well in real-world applications. In practice, we may face many types of attacks where the exact type of adversarial examples in real-world applications can be even unknown. In this paper, motivated by that adversarial examples are more likely to appear near the classification boundary, we study adversarial examples from a new perspective that whether we can defend against adversarial examples by pulling them back to the original clean distribution. We theoretically and empirically verify the existence of defense affine transformations that restore adversarial examples. Relying on this, we learn a defense transformer to counterattack the adversarial examples by parameterizing the affine transformations and exploiting the boundary information of DNNs. Extensive experiments on both toy and real-world datasets demonstrate the effectiveness and generalization of our defense transformer.
\end{abstract}

\section{Introduction}

Deep neural networks (DNNs) have achieved great success in many tasks, such as 
image processing \cite{guo2020zero,moran2020deeplpf}, text analysis \cite{wang2020contournet, zhang2020deep} and speech recognition \cite{ stephenson2019untangling,xu2020discriminative}.
However, DNNs are vulnerable to adversarial examples \cite{liu2019feature, szegedy2013intriguing, zhou2020dast} that are generated from original examples by small perturbations.
This may result in undesirable and even disastrous consequences of DNNs in real-world applications, such as safety accidents in autonomous navigation systems \cite{li2014feature} and medical misdiagnosis \cite{ozbulak2019impact}.
As DNNs are becoming more and more  prevalent, it is necessary to study the problem of how to defend against adversarial examples~\cite{goodfellow2014explaining, liu2019rob, CW_2017, wong2018scaling}. 

\begin{figure*}[t]
	\centering
	{
		\includegraphics[width=0.99\linewidth]{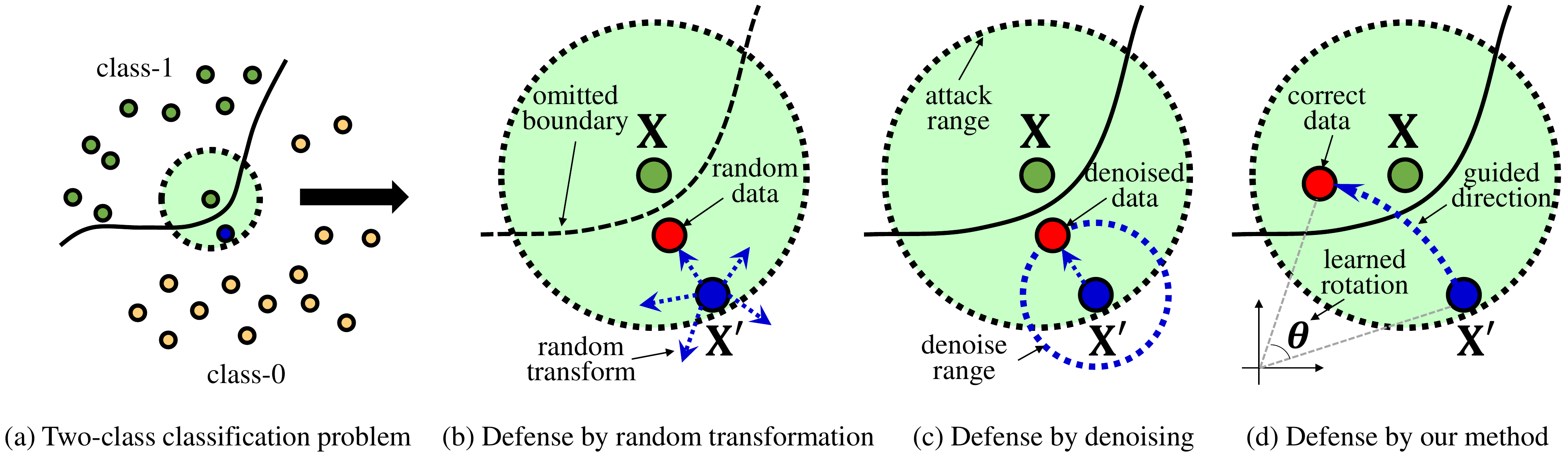}
		\caption{		
			Comparisons of different defense methods, where  
			an adversarial example is generated to cross the classification boundary with a small perturbation.
			Random transformation cannot guarantee to correct the adversarial example without supervision.
			Denoising methods recover the adversarial example in a small range. When the attack range (\eg stAdv or ADef) is larger than the denoise range, these methods may fail to defend.
			Our defense transformer learns a defense affine transformation (\eg rotation) to guide the adversarial example back to the clean data distribution.
		}
		\label{fig:comparision}
	}
\end{figure*}

The key challenge of adversarial defense is how to defend against various types of adversarial attacks, where the exact type of adversarial examples are even unknown in practice. 
Most existing methods, however, consider  some specific types of adversarial examples, making them hard to defend against various adversarial examples raised from real-world applications.
For example, some methods \cite{guo2018countering, raff2019barrage} use a combination of random transformations to defend against adversarial examples.
However, due to the randomness, it is difficult for these methods to transform various types of adversarial examples back to the original distribution of clean data (See Figure~\ref{fig:comparision}(b)).
\tmk{
In addition, the defense methods relying on denoising~\cite{liao2018defense, xie2019feature} seek to handle adversarial examples by removing the introduced noises  (See Figure~\ref{fig:comparision}(c)).
These methods, however, cannot well handle the adversarial examples that are not generated by the same type of noises.
In conclusion, considering the possible various types of unknown attacks in practice, how to learn a general model to defend against various adversarial examples remains an open yet challenging problem.}

In this paper, we address this by \tmk{exploiting} the common characteristics of different types of adversarial examples. 
\tmk{Specifically,} motivated by that adversarial examples are more likely to appear near the classification boundary \cite{zhang2019theoretically}, \blue{we propose to pull adversarial examples back to the original clean distribution.}
We find that there exist defense affine transformations that restore adversarial examples.
Inspired by this, we propose to learn a defense transformer to counterattack adversarial examples by parameterizing the affine transformations. 
In this way, our defense transformer is able to handle various adversarial examples effectively.

We summarize the main contributions as follows: 
\begin{itemize}[leftmargin=*, itemsep=-2pt, topsep=-2pt]
	\item We study the vulnerability of adversarial examples to affine transformations, and theoretically and empirically verify the existence of defense affine transformations that correct adversarial examples.
	\item We propose a defense transformer to counterattack adversarial examples. By parameterizing affine transformations, our defense transformer is able to guide the adversarial examples back to the original distribution of clean data.
	\item  Extensive experiments on toy and real-world datasets demonstrate that our method is able to handle different types of adversarial examples. Also, our method is robust to attacks of different intensities and is able to defend against unseen attacks over different models. 
\end{itemize}

\section{Related Work}

Deep neural networks (DNNs) are vulnerable to adversarial examples \cite{alaifari2018adef,goodfellow2014explaining, madry2017towards, xiao2018spatially} in practice.
To resolve this, many researchers have proposed two main classes of defense methods, including adversarial training methods and defense methods with transformations.

\paragraph{Adversarial training methods.}
These methods \cite{goodfellow2014explaining, madry2017towards, tramer2019adversarial, zhang2019you} seek to defend against adversarial examples by training DNNs in an adversarial manner. 
For example, adversarial logit pairing \cite{kannan2018adversarial} adopts adversarial learning to encourage the predictions of a model on a clean image and its adversarial example to be similar. 
Although these methods improve the robustness of DNNs to adversarial examples, they often perform very limitedly on the clean data \cite{pang2019mixup}.

\begin{figure*}[t]
	\centering
	{
		\includegraphics[width=0.99\linewidth]{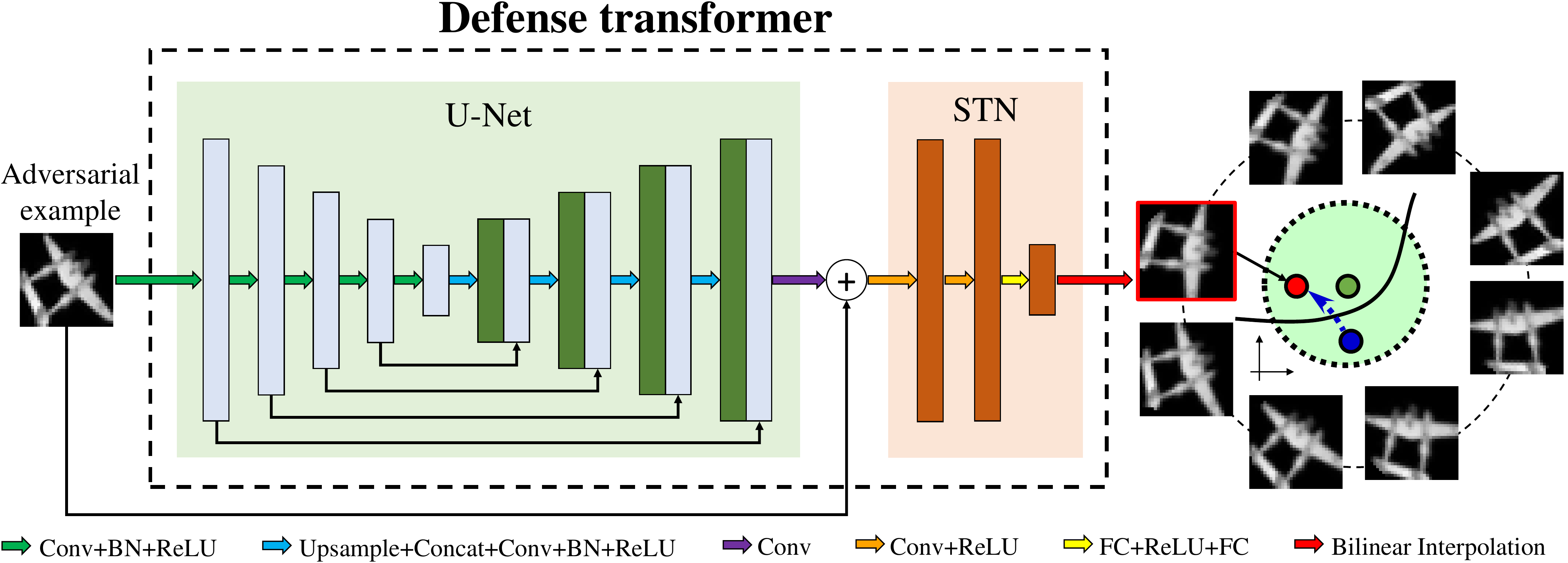}
		\caption{		
			The scheme of the proposed defense transformer. 
			The architecture of the defense transformer consists of a U-Net and a spatial transformer network (STN). 
			\tmk{During} optimization, the defense transformer seeks to learn an affine transformation such that the input adversarial examples can be turned back to the distribution of the clean data.
		}
		\label{fig:framework}
	}
\end{figure*}

\paragraph{Defense methods with transformations.}
These types of methods seek to handle  adversarial examples through transformations.
These methods can be classified into two main categories: random transformation methods and denoising methods.
In random transformation methods, 
cropping-ensemble \cite{guo2018countering} and BaRT \cite{raff2019barrage} 
use a combination of random transformations with different strategies to defend against adversarial examples. 
In addition, the mixup inference method \cite{pang2019mixup} interpolates a sample between an adversarial example and other random clean samples to make predictions.
However, due to the randomness, it is hard for these methods to guide various adversarial examples back to the original distribution of clean data (See Figure \ref{fig:comparision} (b)).
In denoising methods, HGD \cite{liao2018defense} defends against adversarial examples by using high-level representation guided denoiser. Feature denoising \cite{xie2019feature} applies denoising operations directly on features. 
However, it is difficult for these denoising methods  to handle adversarial examples that are not crafted with noises and their attack range is large (See Figure \ref{fig:comparision} (c)).

\section{Problem Definition and Motivations} \label{sec:problem_definition}
\paragraph{Notation.} 
We use calligraphic letters (\eg $\mX$) to denote a space, and use bold lower case letters (\eg $\bx$) to denote a sample (\eg an image).
For a multi-class classification problem, we use $ h $ to denote a classifier.
\zsh{We denote the training dataset by $\mS=\{(\bx_i, y_i)\}_{i=1}^N$, where $\bx_i$ is a sample, and $\by_i$ is the corresponding label.} 
To develop our method, we first define the adversarial example as follows.

\begin{deftn}\label{def:adversarail_example} %\emph{\textbf{(Adversarial example)}}
	Given a pre-trained classifier $h$ and a sample pair $ (\bx, y) $, we call $\widetilde{\bx}$ an adversarial example if there 
	exists a sample $ \widetilde{\bx} {\in} \mX $ such that $ h(\widetilde{\bx}) \neq y $ and $ d(\bx, \widetilde{\bx}) \leq \varepsilon$, where $d(\cdot, \cdot)$ is some distance metric, and $\varepsilon>0$ is a small constant.
\end{deftn}

\ljc{In real-world applications, a DNN model can be attacked by various types of adversarial examples. Based on the types of attacks, they can be classified into two categories: (i) linear adversarial examples (\eg FGSM \cite{goodfellow2014explaining} and PGD \cite{madry2017towards}) have a small perturbation on clean data;
(ii) Non-linear adversarial examples (\eg stAdv \cite{xiao2018spatially} and ADef \cite{alaifari2018adef}) deform clean data by using non-linear functions.
In general, these adversarial examples are more likely to appear near the classification boundary \cite{zhang2019theoretically}. In this sense, they may be vulnerable to some transformations that turn them back to the distribution of clean data. 
Motivated by this, we seek to investigate the vulnerability of adversarial examples.}

\paragraph{Vulnerability of adversarial examples.}
\blue{In practice, adversarial examples are vulnerable to affine transformations \cite{athalye2018synthesizing}.
With the help of this property, existing studies \cite{guo2018countering, raff2019barrage} apply random affine transformations to restore adversarial examples.
However, the random affine transformations are independent of every adversarial example.
To address this,}
we propose to investigate the vulnerability of adversarial examples to affine transformations (\ie rotation, scaling and translation). Formally, the affine transformation (denoted by $f$) can be defined as follows.
\begin{align}
	\begin{bmatrix} 
		u' \\ v' 
	\end{bmatrix}
	= \begin{bmatrix} 
		s\cdot\cos\alpha & -s\cdot\sin\alpha & \delta_u \\ 
		s\cdot\sin\alpha & s\cdot\cos\alpha & \delta_v
	\end{bmatrix} 
	\begin{bmatrix} 
		u \\ v \\ 1 
	\end{bmatrix},
	\label{eqn:affine_transform}
\end{align}
where $[u, v]$ and $[u', v']$ are the coordinates in $\widetilde{\bx}$ and $f(\widetilde{\bx})$, respectively,
$\alpha$ is a rotation angle, $s$ is a scale factor and $\delta_u, \delta_v$ are translation factors.
Based on Eqn. (\ref{eqn:affine_transform}),
we seek to analyze whether a defense affine transformation exists to correct the adversarial example. 
To this end, we use a kernel \wrt Gaussian function \cite{sundaramoorthi2019translation} to develop our analysis.
\begin{thm}
	\emph{\textbf{(Existence of defense affine transformations)}}
	\label{thm:affine_transformation}
	Let $\widetilde{\bx}$ denote an adversarial example, $\bkappa$ denote a convolutional kernel, and $\mP(\cdot)$ be a pooling operator. We define a layer of a classifier as $h_l\left(\widetilde{\bx}\right) = \mP\left( \sigma\left( \bkappa {*} \widetilde{\bx} \right)\right)$, where $*$ is a convolution operation, $\sigma(\cdot)$ is a ReLU activation function. 
	Then, a defense affine transformation $f$ exists if the difference of predictions of the classifier between a transformed sample $f(\widetilde{\bx})$ and a clean data $\bx$ can be small, 
	\begin{align}
		\left\| h_l\left(f\left(\widetilde{\bx}\right) \right) - h_l\left(\bx\right) \right\|_{1} \leq L S \|\bx\|_{\infty} + \|\bz\|_{\infty},
	\end{align}
	where $L$ is a constant \wrt $\bkappa$, $S$ is the image size of $\bx$ and $\bz$ is the difference between $\bx$ and $\widetilde{\bx}$. 
\end{thm}
\vspace{-6pt}
\begin{proof}
See supplementary materials for the proofs.
\end{proof}
\vspace{4pt}

From Theorem \ref{thm:affine_transformation}, a defense affine transformation exists when the upper bound can be sufficiently small.  
In other words, it can correct an adversarial example since the prediction of the classifier on $f(\widetilde{\bx})$ is close to that on $\bx$. 
In this case, we use the defense affine transformations to defend against adversarial examples. However, such a defense affine transformation is non-trivial to obtain in practice. To address this, we define the following defense problem by parameterizing and training affine transformations.

\paragraph{Defense with affine transformations.} 
We study the problem of defense against adversarial examples by transforming adversarial examples to be close to the clean data with affine transformations.
Given a pre-trained classifier $h$, we seek to learn a transformation function $\mT$, such that the prediction of the transformed data $\mT(\widetilde{\bx})$ is close to the label $y$, by solving the following optimization problem:
\begin{align}
	\min\limits_{w} \;\frac{1}{N} \sum\limits_{i=1}^N   \mL\left(h\left(\mT_w\left(\widetilde{\bx}_i\right)\right), y_i\right),
	\label{eqn:problem}
\end{align}
where $\mL$ is some classification loss function, \eg cross-entropy loss, and $w$ is the parameter of $\mT$.
The defense problem has two key challenges. First, how to parameterize and learn a defense affine transformation is non-trivial. 
Second, the types of adversarial examples can be infinite in practice. 
Most existing defense methods using random transformations or denoising  only consider some specific types of adversarial examples and thus may fail to defend against various adversarial examples.

\section{Learning Defense Transformer\tmk{s}}
To find a defense affine transformation for counterattacking various adversarial examples, we propose to learn a novel defense transformer.
To this end, we apply a spatial transformer network (STN) \cite{NIPS2015_5854} to parameterize the affine transformations and develop our defense transformer.
The defense transformer aims to map adversarial examples back to the distribution of clean data.

The overall scheme of our method is shown in Figure \ref{fig:framework}.
To be specific, the defense transformer contains a U-Net \cite{ronneberger2015u} and an STN, where the U-Net uses a shortcut learning to perturb adversarial examples in a proper position, and helps STN to parameterize affine transformations.
Formally, given a coordinate $(u, v)$ in the grid of an adversarial example $\widetilde{\bx}$, then the corresponding coordinate $(u', v')$ of a transformed sample $\mT(\widetilde{\bx})$ can be obtained by a parameterized affine matrix $\bTheta {\in} \mmR^{2\times 3}$, \ie
\begin{align}
	\begin{bmatrix} 
		u' \\ v' 
	\end{bmatrix}
	= \begin{bmatrix} 
		\theta_{11} & \theta_{12} & \theta_{13} \\ 
		\theta_{21} & \theta_{22} & \theta_{23} 
	\end{bmatrix} 
	\begin{bmatrix} 
		u \\ v \\ 1 
	\end{bmatrix},
	\label{eqn:stn}
\end{align}
where $\theta_{ij}$ is the $(i, j)$-th element of the affine matrix $\bTheta$.
In practice, the affine matrix $\bTheta$ can be learned by a localization network $g$ \cite{NIPS2015_5854}, \ie $\bTheta=g(\widetilde{\bx})$.
Then, we use Bilinear interpolation to obtain a transformed sample $\mT(\widetilde{\bx})$.
Note that the defense transformer allows rotation, translation and scale to be applied to an adversarial example.
\blue{Based on the parameterized affine transformation, we next propose to optimize an objective function to defend against different types of adversarial examples.}

\begin{algorithm}[t]
	\caption{Training method of the defense transformer.}
	\label{alg:defense}
	\begin{algorithmic}[1]
		\INPUT 
		Training data $\{(\bx_i, y_i)\}_{i=1}^N$; 
		pre-trained classifier $h$; 
		% 		the number of iterations $T$;
		the type number of attack methods $m$;
		batch size $ n $;
		learning rate $\gamma$;
		hyper-parameter $\lambda$
		\OUTPUT \ljc{Learned} defense transformer $\mT$
		\STATE Apply all types of attack methods to generate a set of adversarial examples $\mA = \{\{(\widetilde{\bx}_i^{(j)}, y_i)\}_{j=1}^{m}\}_{i=1}^N$
		\STATE Initiate the parameter $w$ of the defense transformer 
		\WHILE{not converged}
		% 		\STATE Sample a mini-batch $\mS_n=\{(\bx_i, y_i)\}_{i=1}^n$ from training data
		\STATE Sample a mini-batch $\{\{(\widetilde{\bx}_i^{(j)}, y_i)\}_{j=1}^{m}\}_{i=1}^n$ from the set of adversarial examples 
		\STATE Freeze the classifier $h$ and update the parameter $w$ by gradient descent: \\
		$ w {\leftarrow} w {-}  \gamma \!\!\left\{\! \nabla_{w} \!\!\left[  \frac{1}{mn} \!\!\sum_{i, j} \!\!\left[\!\mL\!\left(\!h\!\left(\!\mT_w\left(\widetilde{\bx}_i^{(j)}\right)\!\right)\!, y_i\!\right) \!\right] \right]  \!\!{+} \lambda {w} \!\right\}$
		\ENDWHILE
	\end{algorithmic}
\end{algorithm}

\paragraph{Learning objective function.}
During the training process, we consider training 
\ljc{a classifier $h$ based on the training dataset $\mS{=}\{(\bx_i, y_i )\}_{i=1}^N$\footnote{We provide the details in the supplemental materials.}. Then, we generate a set of adversarial examples $\mA{=}\{\{(\widetilde{\bx}_i^{(j)}, y_i )\}_{j=1}^m\}_{i=1}^N$ by using $m$ types of attack methods.}
The goal of this paper is to learn a defense transformer by minimizing the following objective function:
\begin{align}\label{eq5}
	\min\limits_{w} \; \frac{1}{mn } \!\!\sum\limits_{i=1}^n \sum\limits_{j=1}^m \!\left[\mL\!\left(h\left(\mT_w\left(\widetilde{\bx}_i^{(j)}\right)\right)\!, y_i\!\right)\! \right] \!\!{+} \frac{1}{2}\lambda\|w\|^2,
\end{align}
where $\mL$ can be any classification loss function, $w$ is the parameter of $\mT$,
\ljc{$\lambda$ is a hyper-parameter, $n$ is the batch size and $n \ll N$.}
\tmk{Here, we use the cross-entropy loss as the loss function and the regularizer to overcome possible overfitting issues.}
Based on Eqn.~(\ref{eq5}), some information about the boundary hidden in the classifier can guide the training of $\mT$.
The detailed training method is shown in Algorithm \ref{alg:defense}.

\begin{figure*}[t]
	\centering
	{
		\includegraphics[width=0.95\linewidth]{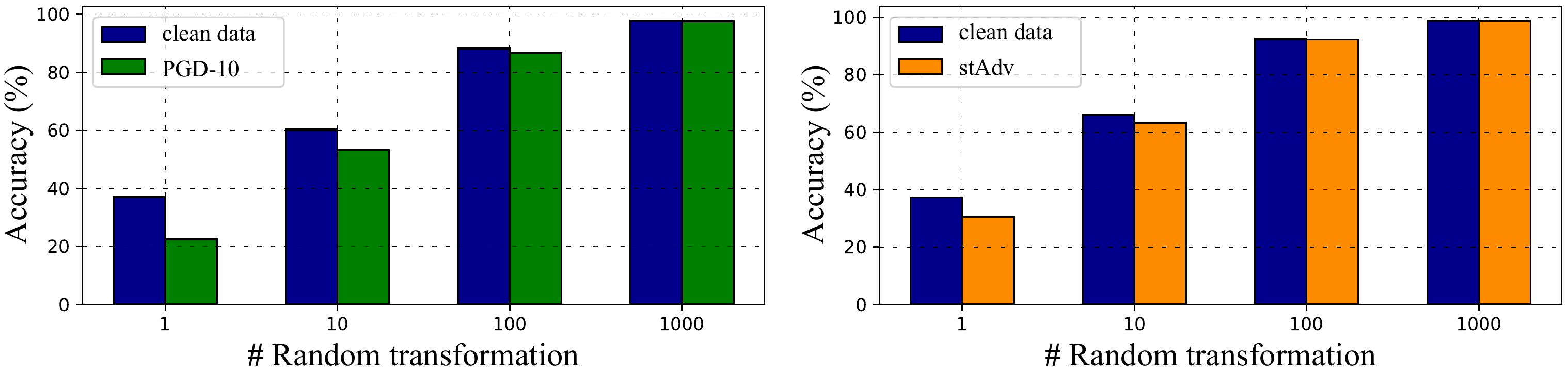}
		\caption{		
			The accuracy of the classifier on clean data and adversarial examples after performing random affine transformations with different maximum numbers  (Left: PGD-10; Right: stAdv) on CIFAR-10. When no affine transformation is applied to clean data, the accuracy for clean data is 93.51\%.
		}
		\label{fig:bar_PGD10_stadv}
	}
\end{figure*}

\begin{figure*}[t]
	\centering
	{
		\includegraphics[width=0.96\linewidth]{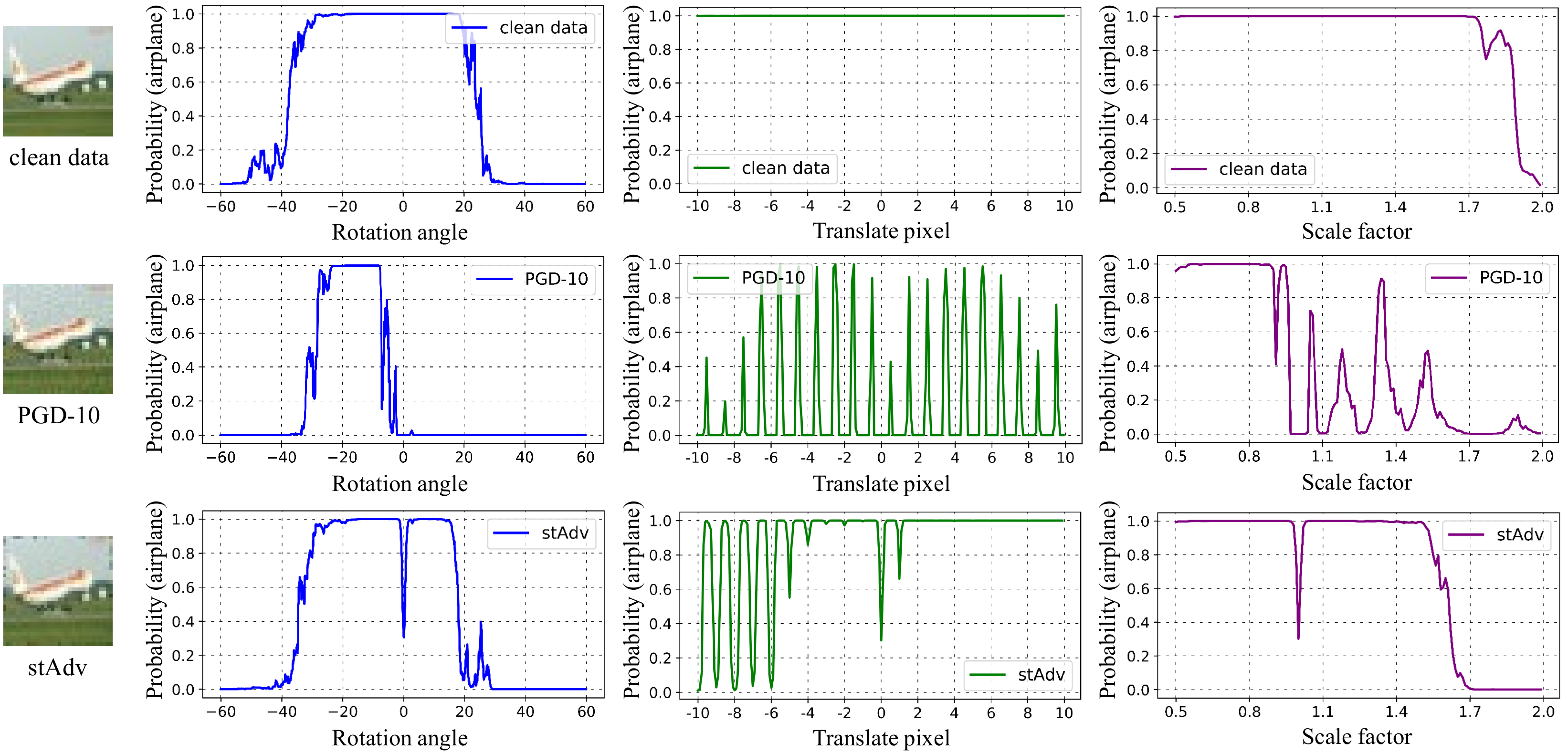}
		\caption{
			Comparisons of different types of data (top: clean data; middle: PGD-10; bottom: stAdv) under different magnitudes of affine transformation (left: rotation; middle: translation; right: scaling).
		}
		\label{fig:sensibility_one_sample}
	}
\end{figure*}

% \newpage
\section{Experiments}

\paragraph{Compared methods.}
For a fair comparison, we compare our method with several state-of-the-art defense methods relying on transformation, including random transformation,
cropping-ensemble \cite{guo2018countering},
high-level representation guided
denoiser (HGD) \cite{liao2018defense},
feature denoising \cite{xie2019feature},
a barrage of random transforms (BaRT) \cite{raff2019barrage} and mixup inference \cite{pang2019mixup}.

\begin{figure*}[t]
	\centering
	{
		\includegraphics[width=1\linewidth]{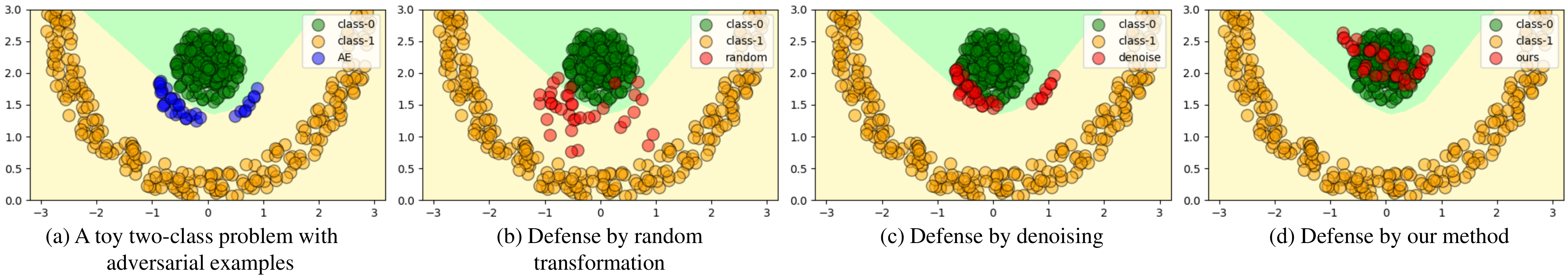}
		\caption{		
			Comparisons of defending against adversarial examples on the toy dataset.
			(a) Given a data distribution with two classes, we generate the adversarial examples (AE) of the class-0 using FGSM. 
			(b) Random transformation can partially recover adversarial examples.
			\ljc{(c) The denoising method can defend against several adversarial examples with a small range.}
			(d) Defense transformer maps all adversarial examples back to the distribution of the correct class.
		}
		\label{fig:toy}
	}
\end{figure*}

\paragraph{Datasets.}
We conduct experiments on a toy dataset and real-world datasets, \ie CIFAR-10, CIFAR-100 \cite{krizhevsky2009learning} and ImageNet \cite{deng2009imagenet}.
In Figure \ref{fig:toy} (a), the toy dataset consists of two classes, represented by orange and green circles. Moreover, we generate adversarial examples (blue) through FGSM \cite{goodfellow2014explaining}. 
More details of the toy dataset are put in the supplemental materials.
On real-world datasets, we generate adversarial examples on the test set based on linear and non-linear attack methods.
Meanwhile, we generate adversarial examples on the training set using multiple linear attack methods.
Linear attack methods include FGSM \cite{goodfellow2014explaining}, I-FGSM \cite{Kurakin2017AdversarialEI}, PGD \cite{madry2017towards}, DeepFool \cite{deepfool_2016_CVPR} and CW-L2 \cite{CW_2017}, where the perturbation levels of these attack methods are provided in the supplementary materials.
Non-linear attack methods  (\eg stAdv \cite{xiao2018spatially} and ADef \cite{alaifari2018adef}) deform images without introducing noises, so they are unsuitable for HGD \cite{liao2018defense}.
\ljc{
For a fair comparison, we do not use non-linear attack methods to generate the training set on CIFAR-10 and CIFAR-100. In addition, we use a small number of attacks (\ie only FGSM) to generate the training set on ImageNet.
More results on the training set with the non-linear attack are put in the supplementary materials.
}

\begin{table*}[t]
	\caption{Performance comparisons (Top-1 accuracy (\%)) of defending against various adversarial examples over ResNet-56 on CIFAR-10 and CIFAR-100. ``AE'' denotes the adversarial examples.}
	\label{tab:resnet56_CIFAR}
	\centering
	\resizebox{1\textwidth}{!}{
		\begin{threeparttable}[b]
			\begin{tabular}{ccccccccclcc}
				\toprule
				\multirow{2}{*}{Datasets} &
				\multirow{2}{*}{Methods} &
				\multirow{2}{*}{Clean} & 
				\multicolumn{6}{c}{Linear AE} & &
				\multicolumn{2}{c}{Non-linear AE} \\
				\cmidrule(r){4-9} \cmidrule(r){11-12} 
				& & & FGSM & I-FGSM & PGD-10 & PGD-100 & DeepFool & CW-L2 & & stAdv & ADef \\
				\midrule
				\multirow{8}{*}{CIFAR-10}
				& Natural training   & \bf93.51   & 21.03   & 0.00  & 0.00  & 0.00  & 5.08  & 3.66  & & 0.39    & 4.25  \\ 
				& Random transformation & 36.98 & 25.21 & 18.08 & 23.39 & 22.11 & 37.10 & 31.06  & & 32.88 & 33.73 \\
				& Cropping-ensemble \cite{guo2018countering}  & 86.31    & 32.56    & 0.46    & 6.47    & 5.12    & 72.89    & 50.08  & & 42.08    & 44.04   \\
				& Mixup inference \cite{pang2019mixup} & 81.47    & 42.38    & 16.58    & 28.40    & 25.89    & 79.51    & 66.31   & & 64.94    & 74.23  \\
				& Feature denoising \cite{xie2019feature}  & 82.32    & 78.68    & 79.13    & 80.48    & 80.77    & 82.00    &80.56  & & 80.06    & 79.90  \\
				& BaRT \cite{raff2019barrage} & 81.48    & 79.53    & 82.15    & 83.35    & 82.91    & 82.27    & 81.01  & & 81.34    & 81.21   \\
				& HGD \cite{liao2018defense}  & 93.23    & 90.24    & 87.79    & 89.27    & 90.51    & 90.21    & 90.21  &  & 44.71    & 82.57  \\
				& Ours & 90.41    & \bf92.34    & \bf89.97    & \bf91.06    & \bf90.59    & \bf90.54    & \bf90.44  &  & \bf85.59  & \bf86.94   \\
				\midrule
				
				\multirow{8}{*}{CIFAR-100}
				& Natural training   & \bf71.08$^*$   & 9.42   & 0.07  & 0.09  & 0.03  & 13.95  & 3.13 &  & 0.68    & 9.88  \\
				& Random transformation & 22.42 & 10.28 & 6.58 & 8.76 & 8.78 & 20.55 & 16.16 &  & 17.44 & 17.86 \\
				& Cropping-ensemble \cite{guo2018countering}     & 57.47   & 13.52   & 0.21  & 1.64  & 0.99  & 48.36  & 33.45 &  & 27.94    & 29.12  \\
				& Mixup inference \cite{pang2019mixup}    & 58.42   & 22.50   & 14.06  & 22.48  & 19.65  & 58.11  & 50.59 &  & 47.08    & 53.95  \\
				& Feature denoising \cite{xie2019feature}     & 51.51   & 48.19   & 47.92  & 49.45  & 49.59  & 51.44  & 49.48 &  & 50.19    & 50.23  \\
				& BaRT \cite{raff2019barrage}    & 50.75   & 47.08   & 47.80  & 49.97  & 49.93  & 51.18  & 49.50 &  & 50.86    & 50.31  \\
				
				& HGD \cite{liao2018defense}    & 70.48   & 59.66   & 46.61  & 57.00  & 58.60  & 61.22  & 65.03 &  & 25.94    & 55.73  \\
				
				& Ours    & 67.22   & \bf75.54$^*$   & \bf63.49  & \bf67.33  & \bf66.55  & \bf66.80  & \bf65.34 &  & \bf59.01    &\bf61.76  \\
				
				\bottomrule
				
			\end{tabular}
			\begin{tablenotes}
				\item $^*$ We will discuss the interesting results separately in Section~\ref{sec:results_cifar10_cifar100}.
			\end{tablenotes}
		\end{threeparttable}
	}
\end{table*}

\paragraph{Implementation details.}
For the defense transformer, the architectures of the U-Net and the spatial transformer network follow \cite{liao2018defense} and \cite{NIPS2015_5854}, respectively. 
In the training process, we adopt an Adam optimizer \cite{kingma2014adam} with $\beta_1=0.5$ and $\beta_2=0.999$, and train the defense transformer with $100$ epochs, a learning rate of $0.001$, \ljc{a hyper-parameter $\lambda=0$} and the batch size of 128. 
In addition, we use the pre-trained ResNet-56 \cite{he2016deep} as the classifier.
\ljc{
More details of the network architecture and the settings of ImageNet are provided in the supplementary materials.
}

\subsection{Existence of Defense Affine Transformations}
To demonstrate the existence of defense affine transformations, we conduct experiments on CIFAR-10, where the linear and non-linear adversarial examples are generated by the PGD-10 and stAdv, respectively. We apply different numbers of random affine transformations  (\ie $1, 10, 10^2, 10^3$)  to both the clean and adversarial examples. For each sample, we will stop transforming if a defense affine transformation is found or the transformation operations have reached the given maximum number.

From Figure~\ref{fig:bar_PGD10_stadv}, we draw two interesting observations. First, when conducting only one random affine transformation, only a small number of adversarial examples are corrected and the accuracy of the clean data decreases a lot. This means that conducting only one random transformation significantly damages the performance of deep models.
Second, with the increasing number of random transformations, the accuracy of the deep model on clean data and adversarial examples goes better and finally performs very well without damage to the clean data. This implies that at least one defense affine transformation exists for defending against adversarial examples and it is difficult to find an affine transformation for defense.

We further investigate the existence of defense affine transformations by applying a transformation (\ie rotation, translation or scaling) with different magnitudes to a sample (more results are put in the supplementary materials). From Figure \ref{fig:sensibility_one_sample}, at certain magnitudes of transformation, the clean and adversarial examples can be correctly classified, which further supports the existence of affine transformations.

\subsection{\tmk{Demonstration} on Toy Dataset}
To investigate the effectiveness of our method, we plot the value surface of the pre-trained classifier in Figure~\ref{fig:toy} on the toy dataset. Here, the value surface depicts the output of the classifier. 
To be specific, the classifier divides two-class training samples into two regions (orange and green).
Taking adversarial examples of class 0 in Figure~\ref{fig:toy} (a) as an example, we show the results in Figure~\ref{fig:toy} (b), (c) and (d).

From Figure ~\ref{fig:toy}, our defense transformer is able to transform all adversarial examples back to the green region, which demonstrates the effectiveness of our method.
In contrast, random transformation fails to turn the most adversarial examples back to the original distribution. 
\ljc{Moreover, the denoising method cannot recover several adversarial examples due to the range of denoise. Thus, it is difficult for random transformation and the denoising method to defend against adversarial examples well.}

\subsection{\ljc{Results on Real-world Datasets}}
\label{sec:results_cifar10_cifar100}
We apply our defense transformer to handle adversarial examples over ResNet-56 and VGG-16-BN and compare the performance with compared methods on CIFAR-10 and CIFAR-100 (See Table \ref{tab:resnet56_CIFAR}).
\ljc{We further verify the effectiveness of our method over ResNet-50, ResNet-101 and DenseNet-161 on ImageNet (See Table \ref{tab:imagenet}). Note that we provide the results over VGG-16-BN in the supplementary materials.}

\paragraph{Results on CIFAR-10.}
From Table \ref{tab:resnet56_CIFAR}, our defense transformer performs the best in handling different types of adversarial examples. 
It implies that the defense transformer is able to improve the model robustness against both linear and non-linear adversarial examples. 
In contrast, defense methods with random transformations perform limitedly since it is hard for these methods to find defense transformations to correct adversarial examples due to randomness. 
For denoising methods, HGD and feature denoising aim to remove noises in the linear adversarial examples or their features and hence cannot perform as well as our method on non-linear adversarial examples.

Note that our defense transformer has greater improvement than random affine transformations on clean data.
When we apply a random affine transformation to clean data, the accuracy for clean data decreases from 93.51\% to 36.98\%. 
In this sense, the incorrect affine transformations seriously degrade the accuracy of clean data. 
Such degraded performance can be alleviated by the defense transformer, although its performance is a bit worse than HGD.

\paragraph{Results on CIFAR-100.} 
We further evaluate our method on a more complex dataset CIFAR-100. In Table \ref{tab:resnet56_CIFAR}, our defense transformer greatly outperforms other defense methods, which suggests that our defense transformer is able to learn defense affine transformations to handle various adversarial examples. 
Moreover, we observe an interesting phenomenon. For the adversarial examples generated by FGSM, our method even outperforms the method without transformation on clean data (75.54\% v.s. 71.08\%). One possible reason is that some misclassified clean samples distribute in the set of adversarial examples of FGSM, and our method transforms them into the distribution of clean data.

\paragraph{Results on ImageNet.}
% \noindent \textbf{Results on ImageNet.}
\ljc{
To further investigate the effectiveness of the proposed method, we compare our defense transformer with HGD over different classifiers (\ie ResNet-50, ResNet-101 and DenseNet-161) on a large-scale dataset ImageNet.
From Table \ref{tab:imagenet}, our defense transformer achieves the best performance than HGD on linear and non-linear adversarial examples. 
Note that we only consider FGSM attack to generate the training set. In this way, our method is able to defend against the unknown adversarial examples, \ie PGD-100 and stAdv.
}

\begin{table}[t]
	\centering 
	\caption{Comparisons (Top-1 accuracy (\%)) of defending against adversarial examples over different classifiers on ImageNet.}
	\label{tab:imagenet}
	\resizebox{0.47\textwidth}{!}{
		\begin{tabular}{ccccc}
			\hline
			Classifier & Method & clean & PGD-100 & stAdv \\
			\hline
			\multirow{2}{*}{ResNet-50}
			& HGD \cite{liao2018defense} & 73.58  & 73.86 & 59.45 \\	
			& Ours & \bf73.59 & \bf74.71 & \bf61.15 \\	
			\hline
			\multirow{2}{*}{ResNet-101}
			& HGD \cite{liao2018defense} & 74.59 & 74.26  & 60.33 \\	
			& Ours & \bf75.37 & \bf74.32 & \bf60.91 \\	
			\hline
			\multirow{2}{*}{DenseNet-161}
			& HGD \cite{liao2018defense} & 74.81 & 74.53 & 59.77\\	
			& Ours & \bf74.94 & \bf74.82 & \bf61.17 \\	
			\hline
		\end{tabular}
	}
\end{table}

\begin{table}
	\centering 
	\caption{Comparisons (Top-1 accuracy (\%)) of defending against unknown adversarial examples over ResNet-56 on CIFAR-10.}
	\label{table:functional}
	\resizebox{0.47\textwidth}{!}{
	\begin{tabular}{cccc}
		\hline
		Method & Functional AE & EOT & One pixel attack\\
		\hline
		HGD \cite{liao2018defense} & 9.00  & 20.72 & 53.59 \\	
		Ours & \bf63.50  &\bf50.71 & \bf64.83 \\	
		\hline
	\end{tabular}
	}
\end{table}

\section{Further Experiments}
 
% \subsection{Comparison with Adversarial Training}
% \ljc{
% In this experiment, we compare our defense transformer with Kurakin adversarial training \cite{Kurakin2017}, Madry adversarial training \cite{madry2017towards} and TRADES \cite{zhang2019theoretically} over WideResNet-34-10 on CIFAR-10.
% Here, we only test our defense transformer on PGD-20 following \cite{zhang2019theoretically} and extract the numbers of adversarial training methods from \cite{zhang2019theoretically} to show in Figure \ref{fig:adversarial_training}.
% From Figure \ref{fig:adversarial_training}, our method achieves the highest accuracy on clean (natural) data and the adversarial examples.
% }

% \begin{figure}[t]
% 	\centering
% 	{
% 		\includegraphics[width=1\linewidth]{adversarial_training_2.pdf}
% 		\caption{
% 			\ljc{Adversarial accuracy \textit{vs.} natural accuracy for adversarial training methods and our defense transformer on CIFAR-10.
% 			}
% 		}
% 		\label{fig:adversarial_training}
% 	}
% \end{figure}

\subsection{Generalization of Defense Transformer}

\ljc{In practice, we often verify the generalization of a pre-trained model on an unseen dataset, \eg test set. Similar to this, we investigate the generalization of defense transformer on unknown adversarial examples (AE) on CIFAR-10.}

\paragraph{\ljc{Generalization on AE from the unknown attacker.}}
\ljc{In this experiment, we compare our defense transformer with baseline methods on more adversarial examples from unknown attack methods, \ie functional adversarial examples \cite{Laidlaw2019}, expectation over transformation (EOT) \cite{athalye2018synthesizing} and one pixel attack\cite{su2019one}.
From Table \ref{table:functional}, our defense transformer outperforms HGD by a large margin. It indicates that our defense transformer is able to generalize to adversarial examples from the unknown attackers.
We provide more results in the supplementary materials.
}

\paragraph{\ljc{Generalization on AE from the unseen classifier.}}
We further investigate the generalization of our defense transformer \ljc{from another perspective}.
To be specific, we first use model A to generate adversarial examples and train the defense transformer. Then, we use the defense transformer to defend against the adversarial examples generated by model B. For simplicity, we define this process as ``A -> B''. We compare our defense transformer with HGD on both linear and non-linear adversarial examples. We report the results of ``ResNet-56 -> B'' in Table~\ref{tab:generalization} and provide more results of ``VGG-16-BN -> B'' in the supplementary materials.

From Table \ref{tab:generalization}, in the process of ``ResNet-56 -> VGG-16-BN'', our defense transformer achieves much better generalization performance than HGD on both linear and non-linear adversarial examples. It implies that our defense transformer is able to defend against unknown adversarial examples generated from the unseen classifier. 
In contrast, the performance of HGD is degraded, especially for non-linear adversarial examples. This indicates that HGD is unable to defend against unseen attacks.

\subsection{Robustness of Defense Transformer}

\ljc{We further evaluate the robustness of our method on adversarial examples (AE) with larger perturbation or generated by some white-box attack methods on CIFAR-100.}

\paragraph{\ljc{Robustness to AE with larger perturbation.}} In this experiment, we verify the robustness of our defense transformer by increasing the perturbation intensities of adversarial examples.
Here, we generate adversarial examples on the test set based on PGD-10 with different perturbation intensities. 
We report the results in Figure \ref{fig:PGD_robustness}, and put more results in the supplementary materials.

From Figure \ref{fig:PGD_robustness}, our defense transformer outperforms HGD under different perturbation intensities. 
Particularly,  when the perturbation intensities become larger, our defense transformer performs much better.
Such an observation verifies that our defense transformer is more robust than HGD.
In contrast, HGD is unable to handle adversarial examples with larger perturbation intensities.
One possible reason is that the perturbation intensity (\ie attack range) may be beyond the denoising range of HGD, as shown in Figure \ref{fig:comparision}.

\begin{table}[t]
	\caption{Comparisons (Top-1 accuracy (\%)) of defending against the adversarial examples generated by unseen classifier.}
	\label{tab:generalization}
	\centering
	\resizebox{0.48\textwidth}{!}{
		\begin{tabular}{cccclcc}
			\toprule
			\multirow{2}{*}{Process} &
			\multirow{2}{*}{Methods} &
			\multicolumn{2}{c}{Linear AE} & &
			\multicolumn{2}{c}{Non-linear AE} \\
			\cmidrule(r){3-4} \cmidrule(r){6-7} 
			& & FGSM & PGD-10 & & stAdv & ADef \\
			\midrule
			
			\multirow{2}{*}{ResNet-56 -> ResNet-56}
			& HGD \cite{liao2018defense}  
			& 90.24 & 89.27  & & 44.71 & 82.57      
			\\
			& Ours
			& \bf92.34 & \bf91.06 & & \bf85.59 & \bf86.94
			\\
			\midrule

			\multirow{2}{*}{ResNet-56 -> VGG-16-BN}
			& HGD \cite{liao2018defense}  
			& 65.06 & 68.96  & & 31.62 & 38.56     \\
			& Ours & \bf76.74 & \bf81.01 &  & \bf80.64 & \bf75.56
			\\
			\bottomrule
			
		\end{tabular}
	}
\end{table}

\begin{table}
	\centering 
	\caption{Comparisons (Top-1 accuracy (\%)) of defending against various adversarial examples (under white-box attacks) over ResNet-56 on CIFAR-100.}
	\label{table:robust}
	\resizebox{0.48\textwidth}{!}{
		\begin{tabular}{ccccc}
			\hline
			Method  
			& PGD-100 ($\epsilon{=}\frac{4}{255}$)
			& PGD-100 ($\epsilon{=}\frac{8}{255}$) 
			& stAdv & ADef \\
			\hline
			HGD \cite{liao2018defense} & 0.06 & 0.03 & 0.69 & 9.52 \\	
			Ours & 0.01 & 0.03 & 1.76 & 11.06 \\
% 			HGD defend against ours' AE & 54.09 & 62.32 \\
% 			Ours defend against HGD's AE & 57.14 & 60.20 \\
			\hline
		\end{tabular}
	}
\end{table}

\begin{table}[t]
	\caption{Comparisons (Top-1 accuracy (\%)) of defending against various adversarial examples over ResNet-56 with different network architectures on CIFAR-10. ``AE'' denotes the adversarial examples.}
	\label{tab:motivations_unet}
	\centering
	\resizebox{0.48\textwidth}{!}{
		\begin{tabular}{ccccclcc}
			\toprule
			\multirow{2}{*}{Method} &
			\multirow{2}{*}{Architecture} &
			\multirow{2}{*}{Clean} & 
			\multicolumn{2}{c}{Linear AE} & &
			\multicolumn{2}{c}{Non-linear AE} \\
			\cmidrule(r){4-5} \cmidrule(r){7-8} 
			& & & FGSM & PGD-10 & & stAdv & ADef \\
			\midrule
			\multirow{3}{*}{Ours}
			& STN   & 78.81    & 57.18    & 64.18  & & 73.76     & 52.89   \\ 
			& Auto-Encoder + STN  & 70.06     & 63.82  & 65.47  & & 67.10     & 67.82  \\ 
			& U-Net + STN   & \bf90.41    & \bf92.34   & \bf91.06  & & \bf85.59      & \bf86.94   \\ 
			\bottomrule
			
		\end{tabular}
	}
\end{table}

\paragraph{\ljc{Robustness to AE under white-box attacks.}}
\ljc{
In this experiment, we discuss the robustness of the defense transformer under white-box attacks. 
Here, the white-box attacks are aware of the exact architecture and the corresponding parameters of a defense method.
Specifically, we apply the PGD-100 attack following \cite{athalye2018robustness} and non-linear attacks (\ie stAdv and ADef) to evaluate the robustness of the defense methods.
From Table \ref{table:robust}, HGD and our method are vulnerable to adversarial examples that are aware of their parameters. 
\blue{Nevertheless, our defense transformer is more robust than HGD since our method is able to defend against more adversarial examples.
It is challenging to defend against white-box attacks.
In the future, it is worth improving the robustness of our method under the white-box attacks.
}
}

\subsection{Ablation Study}
% \ljc{In this section, we discuss more motivations of using U-Net and investigate the impact of hyper-parameter $\lambda$.}

\paragraph{Effectiveness of U-Net.}
We discuss more motivations of using U-Net as a feature representation module.
\ljc{The original spatial transformer network (STN) is very shallow and thus is difficult to learn defense affine transformations for every input data.
To address this, we are motivated to map the original data to a suitable position in the feature space such that a defense transformation can be easier to learn.
To this end, one can apply some representation learning methods (\eg Auto-Encoder and U-Net) to learn feature representations from the original data.}
\ljc{In this experiment, we compare the performance of different architectures in Table \ref{tab:motivations_unet}.
The defense transformer using U-Net and STN achieves the best performance.
We further evaluate the benefit of using U-Net and provide more results in the supplemental materials.}

\begin{figure}[t]
	\centering
	{
		\includegraphics[width=0.86\linewidth]{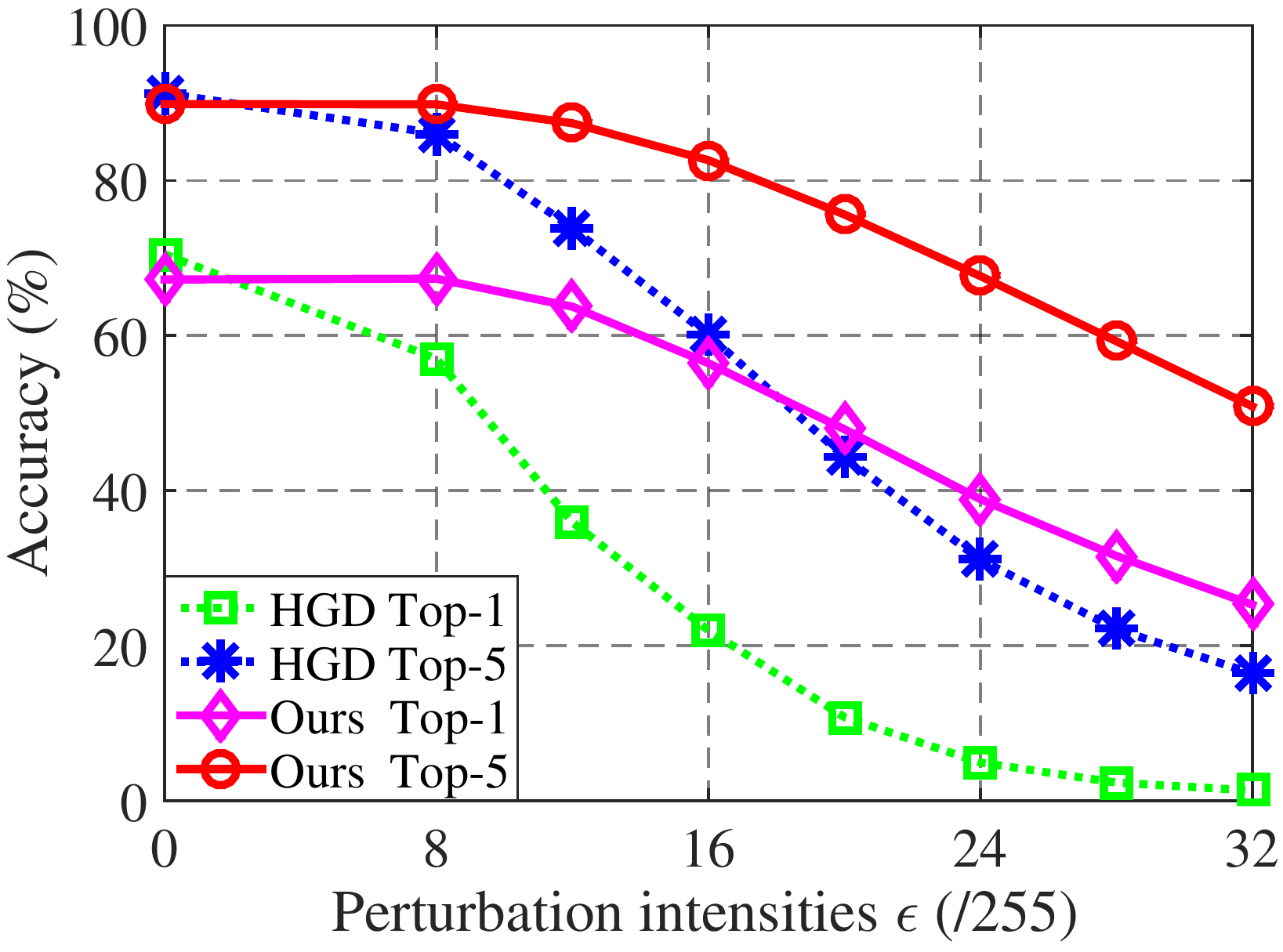}
		\caption{
			\ljc{Robustness \wrt perturbation intensities $\epsilon$. We test our defense transformer and HGD on PGD-10 with different perturbation intensities over ResNet-56 on CIFAR-100.
			}
		}
		\label{fig:PGD_robustness}
	}
\end{figure}

\begin{table}
	\centering 
	\caption{The ablation studies of hyper-parameter $\lambda$ over WideResNet-34-10 on CIFAR-10.}
	\label{table:lambda_impact}
	\resizebox{0.36\textwidth}{!}{
		\begin{tabular}{cccc}
			\hline
			Method	& $\lambda$ & Top-1 Accuracy & Top-5 Accuracy\\
			\hline
			 \multirow{5}{*}{Ours}
			 & 1 & 80.39 & 90.01 \\
			 & 0.1 & 85.90 & 90.09 \\
			 & 0.01 & 88.39 & 90.10 \\
			 & 0.001 & 88.71 & 90.19 \\
			 & 0 & \bf88.91 & \bf90.59 \\
			\hline
		\end{tabular}
	}
\end{table}

\paragraph{Impact of hyper-parameter $\lambda$.} 
\ljc{We further investigate the impact of hyper-parameter $\lambda$ of the regularizer in Eqn.~(\ref{eq5})  following \cite{zhang2019theoretically}. From Table \ref{table:lambda_impact}, the performance of our defense transformer deteriorates when we apply a regularizer into Eqn.~(\ref{eq5}). Thus, we set $\lambda=0$ for all experiments in practice. We provide more details in the supplemental materials.}

\section{Conclusion}
In this paper, we have investigated the vulnerability of adversarial examples and verified that there exist defense affine transformations to turn them back to the distribution of clean data. 
Inspired by this, we learn a defense transformer to defend against different types of adversarial examples.
The proposed defense transformer is robust to the attacks of different intensities and can generalize to defend against unseen attacks over different DNNs.
\blue{More importantly, our defense transformer can be used for data augmentation.}
Extensive experiments on both toy and real-world datasets demonstrate the effectiveness and generalization of our method.

{\flushleft \bf Acknowledgements}. This work was partially supported by Key-Area Research and Development Program of Guangdong Province (2018B010107001, 2019B010155002, 2019B010155001), National Natural Science Foundation of China (NSFC) 61836003 (key project), National Natural Science Foundation of China (NSFC) 62072190, 2017ZT07X183, Tencent AI Lab Rhino-Bird Focused Research Program (No.JR201902), Fundamental Research Funds for the Central Universities D2191240.

{
\small
\bibliographystyle{ieee_fullname} %abbrv
\bibliography{reference}
}

\clearpage

\end{document}